\documentclass[conference]{IEEEtran}
\IEEEoverridecommandlockouts
\usepackage{cite}
\usepackage{amsmath,amssymb,amsfonts}
\usepackage{algorithmic}
\usepackage{graphicx}
\usepackage{textcomp}
\usepackage{xcolor}
\usepackage{siunitx}
\usepackage{tabularx}
\usepackage[bf]{caption}
\usepackage{threeparttable}
\usepackage[singlelinecheck=false, aboveskip=-1pt]{subcaption}

\ifdefined\unit\else
  \ifdefined\NewCommandCopy
    \NewCommandCopy\unit\si
  \else
    \NewDocumentCommand\unit{O{}m}{\si[#1]{#2}}
  \fi
\fi

\def\BibTeX{{\rm B\kern-.05em{\sc i\kern-.025em b}\kern-.08em
    T\kern-.1667em\lower.7ex\hbox{E}\kern-.125emX}}
    
\begin{document}

\title{Steel Phase Kinetics Modeling using\\ Symbolic Regression}

\author{\IEEEauthorblockN{David Piringer}
\IEEEauthorblockA{\textit{Heuristic \& Evolutionary Algorithms Lab} \\
\textit{University of Applied Sciences Upper Austria}\\
4232 Hagenberg, Austria \\
david.piringer@fh-hagenberg.at}
\and
\IEEEauthorblockN{Bernhard Bloder}
\IEEEauthorblockA{\textit{MCL Forschungs GmbH} \\
8700 Leoben, Austria \\
bernhard.bloder@extern.mcl.at}
\and
\IEEEauthorblockN{Gabriel Kronberger}
\IEEEauthorblockA{\textit{Heuristic \& Evolutionary Algorithms Lab} \\
\textit{University of Applied Sciences Upper Austria}\\
4232 Hagenberg, Austria \\
gabriel.kronberger@fh-hagenberg.at}
}

\maketitle

\begin{abstract}
\noindent We describe an approach for empirical modeling of steel phase kinetics based on symbolic regression and genetic programming. The algorithm takes processed data gathered from dilatometer measurements and produces a system of differential equations that models the phase kinetics.
Our initial results demonstrate that the proposed approach allows to identify compact differential equations that fit the data. The model predicts ferrite, pearlite and bainite formation for a single steel type. Martensite is not yet included in the model. Future work shall incorporate martensite and generalize to multiple steel types with different chemical compositions.
\end{abstract}


\section{Introduction}

\noindent Steels are alloys consisting of the main elements iron and carbon and often contain additional elements, which may be deliberate additions for beneficial effects or be present as limited but unavoidable residuals from steelmaking, so called tramp elements. All of these elements are incorporated in the crystalline solid microstructure, which directly affects the properties and performance of the steel~\cite{Krauss2015}. The fact that pure iron has two different crystalline structures at different temperatures allow for steelmaking with great versatility~\cite{Krauss2017}: Austenite at high temperatures (\SI{912}{\degreeCelsius} -- \SI{1394}{\degreeCelsius}) and ferrite at low and room temperatures (\SI{912}{\degreeCelsius}).

\begin{figure}[!h]
\centering
\begin{subfigure}{0.45\linewidth}
  \centering
  \includegraphics[width=0.65\linewidth]{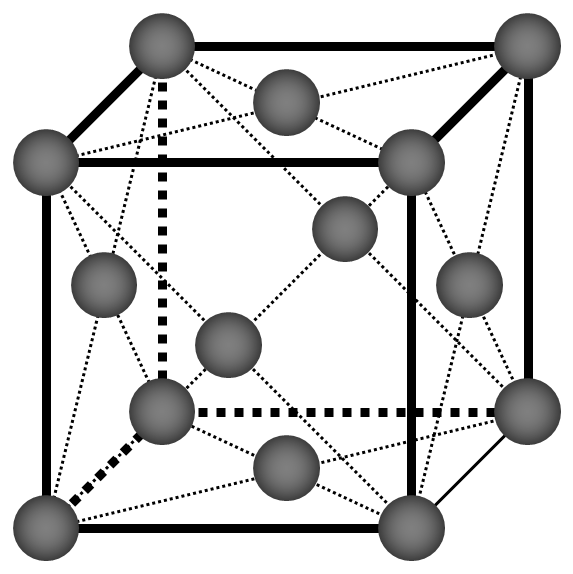}
  \caption{Face-centered cubic structure: Atoms are on each corner and the centers of each face of the cube.}
  \label{fig:fcc}
\end{subfigure}
\hfill
\begin{subfigure}{0.45\linewidth}
  \centering
  \includegraphics[width=0.65\linewidth]{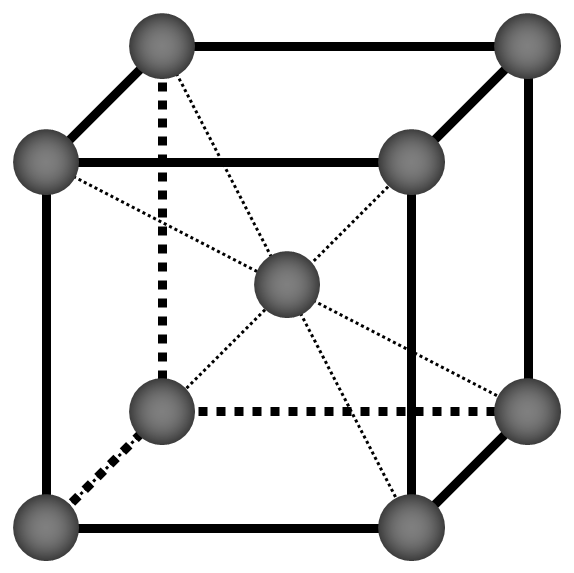}
  \caption{Body-centered cubic structure: Atoms are on each corner and the center of the cube. \\}
  \label{fig:bcc}
\end{subfigure}
\caption{FCC and BCC structures.}
\label{fig:fccBcc}
\end{figure}

The crystal structures of pure iron are shown in Figure \ref{fig:fccBcc}. The shown unit cells define the repeating iron atom arrangements in the crystal. For austenite the structure is face-centered cubic (FCC), for ferrite body-centered cubic (BCC). The FCC structure of austenite is more densely packed than the BCC structure of ferrite, leading to a volume change of approximately 1--3\% for the austenite to ferrite transformation~\cite{Bhadeshia2017}. Dilatometry is a technique that utilizes the change in volume to study phase transformations~\cite{Zhao2007}.

Heat treatment can be used to achieve desired mechanical steel properties such as hardness, strength, ductility or toughness, which are dependent on microstructure and grain size~\cite{Smithells2004}. Steel which is heated to its austenitising temperature may transform to different phases during cooling, each providing a unique combination of properties. Some of these transformations products are (listed in order of their formation with increasing cooling velocity):

\begin{itemize}
	\item Ferrite: BCC crystal structure with other elements in solid solution.
	\item Pearlite: Phase consisting of alternate lamellae of ferrite and cementite (iron carbide, $ Fe_{3}C $), formed on slow cooling.
	\item Bainite: Non-lamellar mixture of ferrite and carbides, which can be classified into upper and lower bainite. Upper bainite consists of plates of cementite in a matrix of ferrite. Lower bainite consists of ferrite needles containing carbide platelets
	\item Martensite: Metastable phase resulting from the diffusionless athermal decomposition of austenite below a certain temperature at sufficiently high cooling rates.
\end{itemize}

\subsection{Transformation model}
\noindent To model the transformation behavior of austenite to its transformation products many models exist, which often rely on the Johnson-Mehl-Avrami-Kolmogorov model (JMAK)~\cite{Johnson1939, Avrami1939, Avrami1940, Avrami1941, Kolmogorov1937}.

The basic equation for isotherm transformation reads as $X = 1 - \exp(-k t^{n})$, where $ X $ is the volume of the transformed phase, and $ k $ and $ n $ are material dependent parameters.

JMAK models may incorporate different nucleation and growth theories and are applicable for the transformation of austenite to ferrite, pearlite and bainite. To describe the transformation to martensite, most commonly the Koistinen-Marburger equation or a variation of it is used~\cite{Koistinen1959}: $X = 1 - \exp(-\alpha (M_{s} - T))$, where $ \alpha $ is a parameter, $ \alpha \approx 0.011 $, $ M_{s} $ is the martensite start temperature (highest temperature at which martensite can form) and $ T $ the temperature of the sample, which is cooled below $ M_{s} $.

Further information on steels with complex microstructures and various modeling approaches can be found in~\cite{Tasan2015}.

\section{Data Acquisition}
\noindent The data is obtained by dilatometer measurements of one steel type. Several measurements with different cooling rates ranging from \SI{0.6}{\unit[per-mode=fraction]{\kelvin\per\second}} to \SI{120}{\unit[per-mode=fraction]{\kelvin\per\second}} were made. 
The data was processed and for the transformation rates, fits for each phase, based on additional information, were calculated such that the overall transformation is described by the transformation of the individual phases.
The result of the data preprocessing are ten data files which contains the cooling rate ($Ks$), the temperature $T$, time $t$, the change rates of volume fractions ($\dot P_1, \dot P_2, \dot P_3, \dot P_4$) and the retained fraction of austenite ($RA$). Because of the high computation time, we decided to use only five data files, which are shown in Figure \ref{fig:phase_kinetics}. We used the combination of the data files to find a model that describes the kinetics for all cooling rates. In this first set of experiments we ignore martensite ($\dot P_4$).

\section{Modeling Approach}
\noindent We used tree-based genetic programming (GP) with a multi-tree representation. Each solution candidate consists of four trees, whereby each tree represents one  expression on the right hand side (RHS) of the system of differential equations (DE).
\begin{align*}
  \ddot P_1 = & f_{P_1}(\dot P_1, \dot P_2, \dot P_3, RA, Ks, T, \theta_{P_1}) \\
  \ddot P_2 = & f_{P_2}(\dot P_1, \dot P_2, \dot P_3, RA, Ks, T, \theta_{P_2}) \\
  \ddot P_3 = & f_{P_3}(\dot P_1, \dot P_2, \dot P_3, RA, Ks, T, \theta_{P_3}) \\
  \dot {RA} = & f_{RA}(\dot P_1, \dot P_2, \dot P_3, RA, Ks, T, \theta_{RA})
\end{align*}
The expression trees are evolved using GP. The algorithm automatically selects  features that are included in each expression. In the same way the functions and operators used in each expression are evolved automatically. We use the same function and terminal set for each expression and apply the same limits for the maximum number of nodes and the maximum depth of each tree. 

The evolved expressions may also include numeric parameters $\theta$ which are memetically optimized as described in~\cite{Kronberger2019}. For fitness evaluation of a solution candidate we first solve the DE system using the measured initial values with a numeric integration scheme (RK45) and then compare to the target values for $\dot P_1, \dot P_2, \dot P_3$ and RA in the data files.

We used all data rows for training to check whether it is in principle possible to evolve a system of DE that fits the measured steel phase kinetics. An analysis of the expected error of the approach with separate training and test sets is left for future work.

\subsection{Algorithm Configuration}
\noindent We used the Age-Layered Population Structure Genetic Algorithm (ALPS-GA), which was invented by Gregory S. Hornby~\cite{10.1145/1143997.1144142}. Similar to the original Genetic Algorithm (GA) by J. Holland~\cite{holland1992adaptation}, is uses the same genetic cycle (with selection, crossover and mutation), but isolated for each age-layer. Each layer represents an age bracket with its own population. The only exception for this isolation is the selection operator, which typically selects parents out of a layer-overlapping mating pool. The default configuration for this mating pool is the current layer (wherein the selection operator is currently working) and the underlying layer with younger individuals. The first layer (youngest age bracket) is regularly reseeding with random individuals (defined by the age gap). New layers open up after a predefined aging scheme, which uses the age gap for calculation. The maximum number of layers is configurable. The age of an individual is defined as follows $1 + \text{Age}_{\text{oldestParent}}$.

For this paper we tried several different parameter settings that produced good results with ALPS-GP for similar problems. A more detailed sensitivity analysis is left for future work. The parameter values are shown in table~\ref{tab:parameter_configuration}. We use and extend the framework HeuristicLab\footnote{https://github.com/heal-research/HeuristicLab}\cite{10.1007/3-211-27389-1_130} for this paper. More details about the assigned operators can be read in~\cite{10.1145/2330784.2330801,affenzeller2009genetic}.

\begin{table}[ht]
\small
\begin{threeparttable}[b]
\begin{tabularx}{\linewidth} {>{\raggedright\arraybackslash}X >{\raggedright\arraybackslash}X }
Parameter & Value \\ 
\hline
Crossover & Subtree Swapping (Probability~=~25\%) \\
Mutator & Multi Symbolic Expression Tree Manipulator \\
Mutation Probability & 10\% \\
Elites & 1 \\
Population Size (per Layer) & 200 \\
Selector & Generalized Rank (Pressure~=~5) \\ 
Mating Pool Range & 1 (Current + Underlying) \\ 
Age Gap & 8 \\ 
Aging Scheme & Polynomial \\
Age Layers & 16 \\ 
max. Generations & 2000 \\ 
Terminal Set & State variables and real-valued parameters \\
Function Set & $+, \times, \div, x^2, \exp, \tanh$, analytic quotient\tnote{1} \\
\hline
\end{tabularx}
\begin{tablenotes}
\item [1] $AQ(x,y) = x/\sqrt{1+y^2}$
\end{tablenotes}
\end{threeparttable}
\caption{Parameter configuration for the ALPS-GA.}
\label{tab:parameter_configuration}
\end{table}

\section{Results}
\begin{figure*}
    \captionsetup[subfigure]{justification=centering}
    \centering
    \begin{subfigure}[t]{0.49\linewidth}
        \centering
        \caption{Cooling rate \SI{0.6}{\unit[per-mode=fraction]{\kelvin\per\second}}}
        \label{fig:plot_0_6}
        \includegraphics[width=\linewidth]{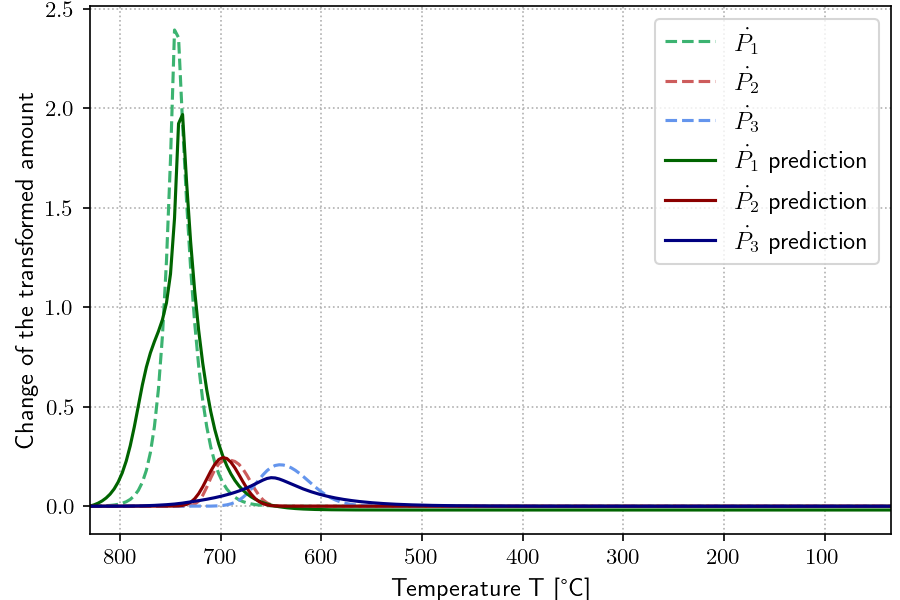}
    \end{subfigure}
    \begin{subfigure}[t]{0.49\linewidth}
        \centering
        \caption{Cooling rate \SI{2.5}{\unit[per-mode=fraction]{\kelvin\per\second}}}
        \label{fig:plot_2_5}
        \includegraphics[width=\linewidth]{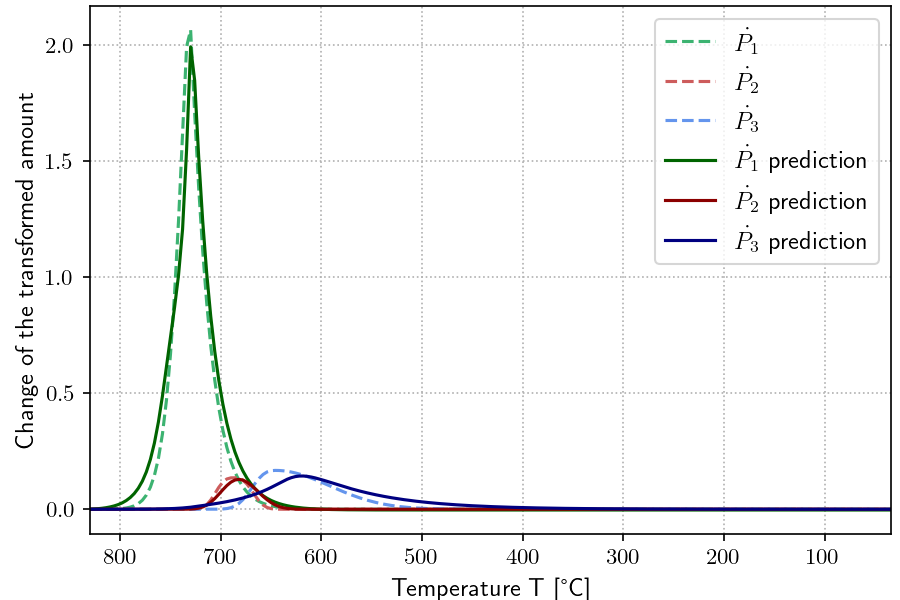}
    \end{subfigure}\vspace{4pt}
    \hfill
    \begin{subfigure}[t]{0.49\linewidth}
        \centering
        \caption{Cooling rate \SI{10}{\unit[per-mode=fraction]{\kelvin\per\second}}}
        \label{fig:plot_10_0}
        \includegraphics[width=\linewidth]{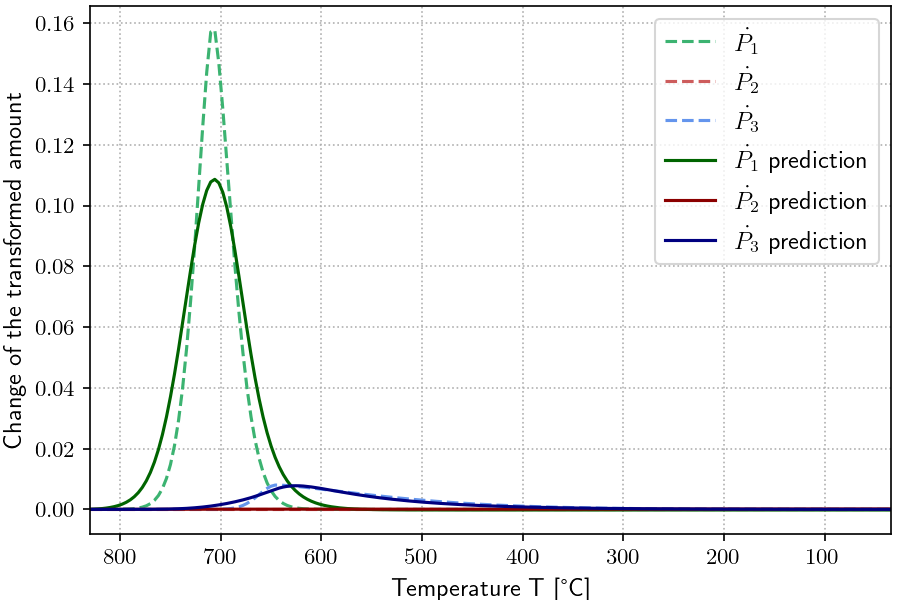}
    \end{subfigure}
    \begin{subfigure}[t]{0.49\linewidth}
        \centering
        \caption{Cooling rate \SI{40}{\unit[per-mode=fraction]{\kelvin\per\second}}}
        \label{fig:plot_40_0}
        \includegraphics[width=\linewidth]{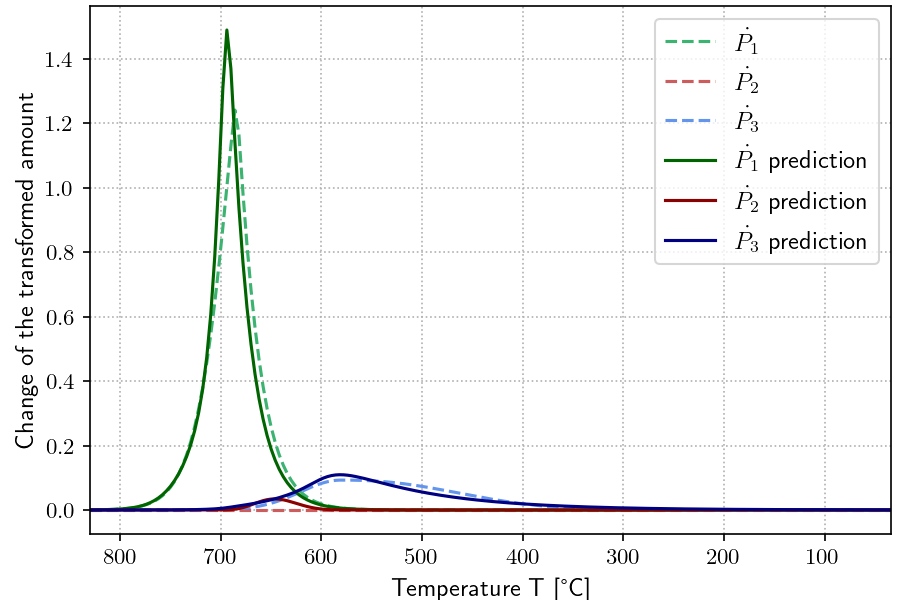}
    \end{subfigure}\vspace{4pt}
    \hfill
    \begin{subfigure}[t]{0.49\linewidth}
        \centering
        \caption{Cooling rate \SI{80}{\unit[per-mode=fraction]{\kelvin\per\second}}}
        \label{fig:plot_80_0}
        \includegraphics[width=\linewidth]{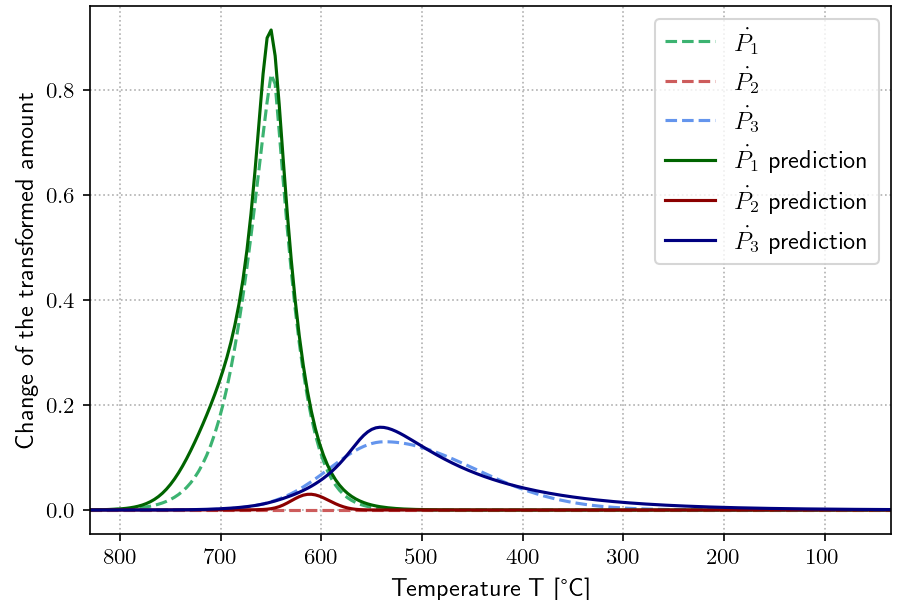}
    \end{subfigure}
    \caption{Data and predictions (training) for five cooling rates. Each plot shows the rate of change of the volume fraction for $\dot{P_1}$ (ferrite), $\dot{P_2}$ (pearlite) and $\dot{P_3}$ (bainite).}
    \label{fig:phase_kinetics}
\end{figure*}

\noindent Figure~\ref{fig:phase_kinetics} shows data and predictions for five different cooling rates (in \unit[per-mode=symbol]{\kelvin\per\second}). Because we use all data rows for training, all predictions reflect the training set. Each panel in the figure shows the change rate of the transformed amount over temperature $T$, where $T$ ranges from \SI{830}{\degreeCelsius} down to \SI{34}{\degreeCelsius}, for  $\dot{P_1}$ (ferrite), $\dot{P_2}$ (pearlite) and $\dot{P_3}$ (bainite) as well as their predictions. At the current stage, the amount of $\dot P_4$ (martensite) is ignored.

\subsection{Models}
\noindent The equations \ref{eq:phase1} (phase 1), \ref{eq:phase2} (phase 2), \ref{eq:phase3} (phase 3) and \ref{eq:restaustenite} (retained austenite) are parts of one solution, which represent the best result found in terms of the normalized mean squared error (NMSE). Every equation contains numeric parameters (tagged with $c_x$). 

\begin{center}
\vspace{-8px} 
\begin{equation}
\label{eq:phase1}
\begin{aligned}
\ddot{P_1} ={} & \left( AQ(c_0, \tanh(c_1 \, RA)) + c_2 \right) \, \\ 
         & \left( c_3 \, \dot{P_1} + AQ(c_4, (c_5 \, Ks)^2) \right) \, 
         c_6 + c_7
\end{aligned} 
\end{equation}  
\begin{equation}
\scriptsize
\begin{aligned}
\nonumber
c_{0} &= 3.591 & c_{3} &= 6.4351 \cdot 10^{-2} & c_{6} &= 5.1832 \\
c_{1} &= 2.2151 & c_{4} &= 4.2717 & c_{7} &= 5.0527 \cdot 10^{-4} \\
c_{2} &= -3.031 & c_{5} &= 56.255
\end{aligned}  
\end{equation} 
\end{center}

\begin{center}
\vspace{-4px} 
\begin{equation}
\label{eq:phase2}
\begin{aligned}
\ddot{P_2} ={} ( & c_{0} \, \dot{P_1} \, ( c_{0} \, \dot{P_2} + c_{1} ) \, c_{2} + 
 c_{3} \, \dot{P_3} \, c_{3} \, \dot{P_2} \, c_{4} + c_{5} 
) \, c_{6} + c_{7}
\end{aligned}  
\end{equation}  
\begin{equation}
\scriptsize
\begin{aligned}
\nonumber
c_{0} &= 0.96475 & c_{3} &= 1.0441 & c_{6} &= 0.65098 \\
c_{1} &= -7.6099 \cdot 10^{-5} & c_{4} &= -3.9235 & c_{7} &= 2.0752 \cdot 10^{-7} \\
c_{2} &= 0.68128 & c_{5} &= 2.5571 \cdot 10^{-5}
\end{aligned}  
\end{equation}  
\end{center}

\begin{center}
\vspace{-4px} 
\begin{equation}
\label{eq:phase3}
\begin{aligned}
\ddot{P_3} ={}  ( & c_{0}  \, \dot{P_1} + c_{1}\, \dot{P_3} \, 
\tanh ( \tanh ( c_{2} \, \dot{P_1} + c_{3} )) \, c_{4} 
)  \, c_{5} + c_{6}
\end{aligned}  
\end{equation}  
\begin{equation}
\scriptsize
\begin{aligned}
\nonumber
c_{0} &= 3.1903 \cdot 10^{-4} & c_{3} &= 0.75539 & c_{6} &= 1.4733 \cdot 10^{-5} \\
c_{1} &= 0.99678 & c_{4} &= -9.0097 \cdot 10^{-2} \\
c_{2} &= -406.1 & c_{5} &= 0.63386
\end{aligned}  
\end{equation}  
\end{center}

\begin{center}
\vspace{-4px} 
\begin{equation}
\label{eq:restaustenite}
\begin{aligned}
\dot{RA} =  ( &  AQ(c_0, c_{1}  \, RA + c_{2}) + \\ 
& c_{3} \, \dot{P_1} \,  ( AQ(c_4, c_{5}  \, \dot{P_1}) + c_{6} )  + c_{7} 
) \, c_{8} + c_{9}
\end{aligned}  
\end{equation}  
\begin{equation}
\scriptsize
\begin{aligned}
\nonumber
c_{0} &= 0.50793 & c_{4} &= -1.5443 \cdot 10^{-3} & c_{7} &= -0.28694 \\
c_{1} &= 9.0808 & c_{5} &= 0.7967 & c_{8} &= 6.6264 \cdot 10^{-2} \\
c_{2} &= -3.368 & c_{6} &= 0.12045 & c_{9} &= 1.3127 \cdot 10^{-2} \\
c_{3} &= 0.91246
\end{aligned}  
\end{equation} 
\end{center}

\section{Discussion}
\noindent The results of our approach are promising and indicate the possibility to use GP with a multi-tree representation for DE systems. Despite the strong simplification, we are optimistic to find good and suitable solutions for the given problem formulation of steel phase kinetics modeling. The found formulas are compact and readable. Additionally, each formula contains self-references or references to other formulas, which is preferred and necessary for a functional DE system. The learned system fits the targets quite well but still has some margin for improvement. 
The high computational cost of the current implementation has to be considered, as it makes it difficult to experiment with a broader spectrum of configurations.
Nevertheless, there are still improvements and extensive experiments planned for future work. Martensite should be incorporated into the model and the model must be generalized to multiple steel types. For this it could be useful to include the grain size as an additional parameter.
Furthermore, it is necessary to split the data set into training and test sets to make sure the model is not overfitting.
A severe limitation of the current approach is that the model uses estimated values for the volume fractions which were fit using additional information. These estimates might be incorrect and lead to an inaccurate model. It remains an open question how we can learn a model for the kinetics solely from the dilatometer data.

\section*{Acknowledgements}
\small
\noindent The authors gratefully acknowledge the financial support under the scope of the COMET program within the K2 Center “Integrated Computational Material, Process and Product Engineering (IC-MPPE)” (Project No 886385). This program is supported by the Austrian Federal Ministries for Climate Action, Environment, Energy, Mobility, Innovation and Technology (BMK) and for Digital and Economic Affairs (BMDW), represented by the Austrian Research Promotion Agency (FFG), and the federal states of Styria, Upper Austria and Tyrol.

\bibliographystyle{ieeetr}
\bibliography{references}

\begin{thebibliography}{10}

\bibitem{Krauss2015}
G.~Krauss, {\em Steels: Processing, Structure, and Performance}.
\newblock {ASM} International, Jan. 2015.

\bibitem{Krauss2017}
G.~Krauss, ``Physical metallurgy of steels,'' in {\em Automotive Steels},
  pp.~95--111, Elsevier, 2017.

\bibitem{Bhadeshia2017}
H.~Bhadeshia and R.~Honeycombe, {\em Steels: Microstructure and Properties}.
\newblock Elsevier Science and Technology, Jan. 2017.

\bibitem{Zhao2007}
J.-C. Zhao, ed., {\em Methods for Phase Diagram Determination}.
\newblock Elsevier Science, 2007.

\bibitem{Smithells2004}
W.~F. Gale and T.~C. Totemeier, eds., {\em Smithells Metals Reference Book}.
\newblock Amsterdam Boston: Elsevier Butterworth-Heinemann, 2004.

\bibitem{Johnson1939}
W.~A. Johnson and R.~F. Mehl, ``Reaction kinetics in processes of nucleation
  and growth,'' {\em Trans. Am. Inst. Min. Metall. Eng.}, vol.~135,
  pp.~416--442, 1939.

\bibitem{Avrami1939}
M.~Avrami, ``Kinetics of phase change. {I} general theory,'' {\em The Journal
  of Chemical Physics}, vol.~7, pp.~1103--1112, Dec. 1939.

\bibitem{Avrami1940}
M.~Avrami, ``Kinetics of phase change. {II} transformation-time relations for
  random distribution of nuclei,'' {\em The Journal of Chemical Physics},
  vol.~8, pp.~212--224, Feb. 1940.

\bibitem{Avrami1941}
M.~Avrami, ``Granulation, phase change, and microstructure kinetics of phase
  change. {III},'' {\em The Journal of Chemical Physics}, vol.~9, pp.~177--184,
  Feb. 1941.

\bibitem{Kolmogorov1937}
A.~Kolmogorov, ``A statistical theory for the recrystallisation of metals,''
  {\em Izvestiya Akademii Nauk SSSR}, vol.~3, 1937.

\bibitem{Koistinen1959}
D.~Koistinen and R.~Marburger, ``A general equation prescribing the extent of
  the austenite-martensite transformation in pure iron-carbon alloys and plain
  carbon steels,'' {\em Acta Metallurgica}, vol.~7, pp.~59--60, Jan. 1959.

\bibitem{Tasan2015}
C.~Tasan, M.~Diehl, D.~Yan, M.~Bechtold, F.~Roters, L.~Schemmann, C.~Zheng,
  N.~Peranio, D.~Ponge, M.~Koyama, K.~Tsuzaki, and D.~Raabe, ``An overview of
  dual-phase steels: Advances in microstructure-oriented processing and
  micromechanically guided design,'' {\em Annual Review of Materials Research},
  vol.~45, pp.~391--431, Jul. 2015.

\bibitem{Kronberger2019}
G.~Kronberger, L.~Kammerer, and M.~Kommenda, ``Identification of dynamical
  systems using symbolic regression,'' in {\em Computer Aided Systems Theory
  – EUROCAST 2019: 17th International Conference, Las Palmas de Gran Canaria,
  Spain, February 17–22, 2019, Revised Selected Papers, Part I}, (Berlin,
  Heidelberg), p.~370–377, Springer-Verlag, 2019.

\bibitem{10.1145/1143997.1144142}
G.~S. Hornby, ``{ALPS}: The age-layered population structure for reducing the
  problem of premature convergence,'' in {\em Proceedings of the 8th Annual
  Conference on Genetic and Evolutionary Computation}, GECCO '06, (New York,
  NY, USA), p.~815–822, Association for Computing Machinery, 2006.

\bibitem{holland1992adaptation}
J.~H. Holland, {\em Adaptation in natural and artificial systems: an
  introductory analysis with applications to biology, control, and artificial
  intelligence}.
\newblock MIT press, 1992.

\bibitem{10.1007/3-211-27389-1_130}
S.~Wagner and M.~Affenzeller, ``{H}euristic{L}ab: A generic and extensible
  optimization environment,'' in {\em Adaptive and Natural Computing
  Algorithms} (B.~Ribeiro, R.~F. Albrecht, A.~Dobnikar, D.~W. Pearson, and
  N.~C. Steele, eds.), (Vienna), pp.~538--541, Springer Vienna, 2005.

\bibitem{10.1145/2330784.2330801}
M.~Kommenda, G.~Kronberger, S.~Wagner, S.~Winkler, and M.~Affenzeller, ``On the
  architecture and implementation of tree-based genetic programming in
  heuristic{L}ab,'' in {\em Proceedings of the 14th Annual Conference Companion
  on Genetic and Evolutionary Computation}, GECCO '12, (New York, NY, USA),
  p.~101–108, Association for Computing Machinery, 2012.

\bibitem{affenzeller2009genetic}
M.~Affenzeller, S.~Wagner, S.~Winkler, and A.~Beham, {\em Genetic algorithms
  and genetic programming: modern concepts and practical applications}.
\newblock Chapman and Hall/CRC, 2009.

\end{thebibliography}
\end{document}